\documentclass[10pt,twocolumn,letterpaper]{article}

\usepackage{times}
\usepackage{epsfig}
\usepackage{graphicx}
\usepackage{amsmath}
\usepackage{amssymb}
\usepackage{comment}
\usepackage{url}
\usepackage{authblk}

\begin{document}

\title{GPU-accelerated real-time stixel computation}


\author[1]{D. Hernandez-Juarez}
\author[1]{A. Espinosa}
\author[2]{D. V{\'a}zquez}
\author[2]{A. M. L{\'o}pez}
\author[1]{\\J. C. Moure}
\affil[1]{Computer Architecture \& Operating Systems Department (CAOS) at Universitat Autonoma de Barcelona}
\affil[2]{Computer Vision Center}
\renewcommand\Affilfont{\itshape\small}

\maketitle

\begin{abstract}
The Stixel World is a medium-level, compact representation of road scenes that abstracts millions of disparity pixels into hundreds or thousands of stixels. The goal of this work is to implement and evaluate a complete multi-stixel estimation pipeline on an embedded, energy-efficient, GPU-accelerated device. This work presents a full GPU-accelerated implementation of stixel estimation that produces reliable results at 26 frames per second (real-time) on the Tegra X1 for disparity images of $1024$$\times$$440$ pixels and stixel widths of 5 pixels, and achieves more than 400 frames per second on a high-end Titan X GPU card.
\end{abstract}

\section{Introduction}

Advanced driver assistance systems (ADAS), autonomous vehicles, robots and other intelligent devices can estimate the distance of objects and the free space in a given scene by computing depth information from stereo camera systems or LIDARs. The large amount of low-level per-pixel depth data is very costly to process and commonly a medium-level representation known as the stixel world \cite{Badino2009} is used for road scenes. It relies on the fact that man-made environments mostly present horizontal and vertical planar surfaces, like roads, sidewalks or soil (horizontal), and buildings, pedestrians or cars (vertical).

\begin{figure*}
  \includegraphics[width=\textwidth]{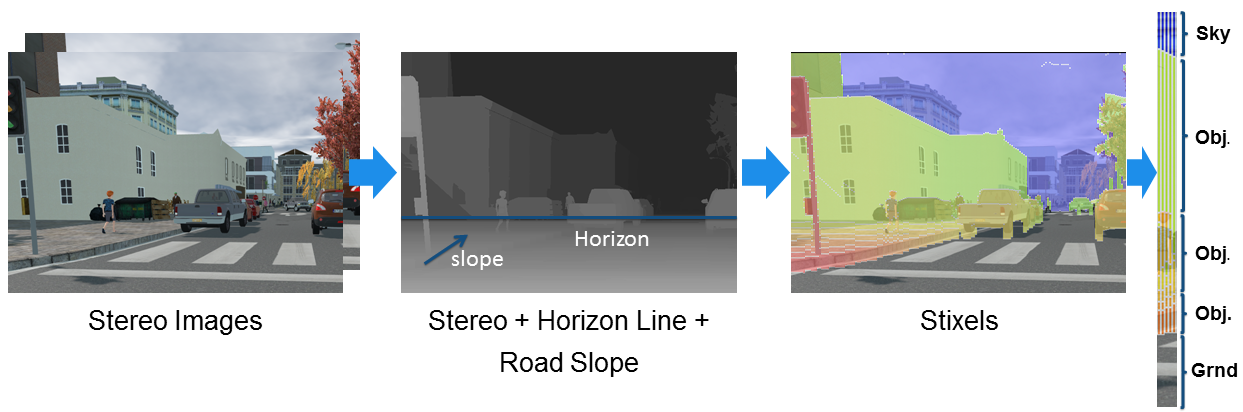}
  \caption{Stixel world: taking the dense disparity map as the input, and estimating a certain ground slope and horizon line, columns are segmented into \emph{stixels} and classified into ground, object and sky categories.}
  \label{fig:overview}
\end{figure*}

Stixels are segments of image columns that represent obstacles. They provide a compact representation that converts millions of disparity pixels to hundreds or thousands of stixels. Pfeiffer and Franke \cite{Pfeiffer2011} proposed an extended representation that allows multiple stixels per column, providing a richer representation of the scene (See Fig. \ref{fig:overview}). Stixels are the basis for multiple extensions such as tracking \cite{pfeiffer_iv10}, grouping \cite{Enzweiler2012} or semantics \cite{Schneider2016}, and also serves as the input for further processing, like pedestrian detection \cite{Benenson2012}.

Calculating stixels is a complex task, comparable to that of generating dense stereo information. As a consequence, the algorithm implemented on a multi-core CPU by \cite{PfeifferThesis} does not fulfill the real-time nor the energy-efficiency requirements of autonomous driving applications. Dedicated hardware designs (e.g. FPGA or ASIC) may achieve these goals, but are very inflexible and costly regarding changes in the algorithms, like combining stixels and semantic segmentation \cite{Schneider2016}. We explore GPU acceleration as an alternative.

The appearance of embedded GPU-accelerated systems, like the NVIDIA Jetson TX1 and DrivePX platforms, opens the door for low-cost and low-energy consumption, real-time stixel computation. GPUs are very well suited for algorithms exhibiting massive and embarrassing parallelism, but may suffer high performance inefficiencies with algorithms that contain inherent dependencies, as those using dynamic programming techniques like the stixel algorithm presented by \cite{Pfeiffer2011}. Careful work distribution and task cooperation, coupled with an appropriate data layout design, may overcome those difficulties and achieve competitive performance. Recently, \cite{HCE2016} proved that GPU-acceleration of dense stereo computation using Semi Global Matching -mostly based on a 1D dynamic programming algorithm- can be successfully achieved. 

The objective of this work is to implement and evaluate a GPU-accelerated software implementation\footnotemark of a complete multi-stixel estimation pipeline, a missing work up to the best of our knowledge. We discuss the optimized massively parallel schemes and data layouts of each of the algorithms involved. Our proposal runs on a single Tegra X1 chip almost two times faster (26 fps versus 13.3 fps) and achieves 25 times better performance per watt ratio than the multicore implementation in \cite{PfeifferThesis}, for the same disparity image size (1024$\times$440 px) and stixel width (5 px), providing the same high-quality results. The proposed design achieves 413 fps on a high-end Titan X GPU card, more than 30 times faster than \cite{PfeifferThesis} with a similar energy envelope.
\footnotetext{https://github.com/dhernandez0/stixels}

The remainder of this paper is structured as follows. Section \ref{sec:related_work} reviews the state of the art on stixel computation. Section \ref{sec:stixels} reviews stixel formulation. Section \ref{sec:gpu} explains general basic concepts of GPU optimization while section \ref{sec:parallelization} explains our proposed GPU-based optimizations for real-time stixel computation. Finally, in section \ref{sec:results} the accuracy and performance of our proposed method is evaluated.

\section{Related work}
\label{sec:related_work}


The stixel world was introduced by Badino et al. \cite{Badino2009} as an intermediate representation suitable for high-level tasks such as object detection or tracking \cite{Benenson2012, pfeiffer_iv10}. The ground surface of a scene, also called free-space (space without obstacles), is estimated using \cite{Wedel2009} from a depth map computed with an stereo algorithm such as SGM \cite{Hirschmuller2008}. Then, dynamic programming is employed in order to find the bottom of the stixels and their heights.

In order to achieve real-time stixel estimation on CPU, Benenson et al. \cite{Benenson2011} developed a less accurate method to calculate the ground surface by accumulating matching costs into vertical-disparity space, and then computing the bottom and height of stixels in disparity-cost space. Disparities are computed only for object stixels and not for ground.

The work from \cite{Badino2009, Benenson2011} considers only a single stixel per column, which represents an incomplete world model that e.g. cannot represent a pedestrian and a building on the same column. Pfeiffer and Franke \cite{Pfeiffer2011} developed a unified probabilistic approach, also solved with dynamic programming, that considers the occurrence of multiple stixels per column.

The stixel world is the fundamental building block for more informative representations. Some extensions are stixel tracking, which provides a vector of velocity for each stixel \cite{pfeiffer_iv10}; stixel grouping, where stixels that seem to belong to the same object are grouped together \cite{Enzweiler2012}; and, finally, semantic stixels represent a combination of the stixel world and semantic segmentation \cite{Schneider2016}. All these extensions would greatly benefit from real-time stixel computation.

Muffert et al. \cite{Muffert} claim to run their FPGA implementation at 25 fps with stixel widths of 5 px, but the authors do not indicate the image resolution.

\section{The Stixel World}
\label{sec:stixels}
Stixels provide a compact representation of man-made scenes, mainly roads, modeled as horizontal and vertical surfaces, which identify the ground and the objects on the scene, such as buildings, pedestrians, cars, or traffic lights. We follow the stixel world model defined in \cite{PfeifferThesis, Pfeiffer2011}, where the reader can find more details.

Stixels are segments of image columns, with a certain height and distance, that are classified as ground, object or sky. Fig. \ref{fig:overview} shows an example with a detailed column that is segmented into 5 different stixels. A comprehensive example can be found at the end of the paper, in Fig. \ref{fig:stixels_example}.

A hard assumption of the model is that stixels in different columns are independent. The ground slope and horizon line are assumed to be known for each image. The stixel computation problem is addressed using a unified probabilistic approach that incorporates real-world constraints such as perspective ordering, and is formulated as a MAP estimation problem that delivers an optimal segmentation of the columns with respect to the free-space and obstacle information \cite{Pfeiffer2011}. A Dynamic Programming (DP) solving scheme incorporates the prior knowledge in order to minimize the global cost of the solution.

The following constraints are modeled as prior probabilities:
(1) \textit{bayesian information criterion (BIC)}: there is a small number of objects in the scene; (2) \textit{gravity constraint}: flying objects are unusual; (3) \textit{ordering constraint}: the upper of two staggered stixels classified as objects is expected to be further away; (4) \textit{staggering constraint}: some configurations, like having a ground stixel above a sky stixel, are impossible; and (5) \textit{diving constraint}: the base point depth of an object stixel should be equal or greater than the corresponding ground depth.

Next we will formalize the problem, define the DP recursive equation, and discuss some algorithmic optimizations.

\subsection{Formal description of MAP problem}

The original disparity image $D$ of $w$$\times$$h$ pixels is pre-processed to convert each column of width $s$ into a column of a single pixel, resulting in a reduced disparity image $D^{'}$ of $\frac{w}{s}$$\times$$h$ pixels. Since all columns are processed independently, we restrict our description to finding the optimal stixel segmentation of a single column $D_i$.

A stixel is a segment $s_n = \{v_{n}^{b}, v_{n}^{t}, c_n\}$, where $v_{n}^{b}$ and $v_{n}^{t}$ are, respectively, the base (beginning) and top (ending) positions in the column ($0$$\leq$$v_{n}^{b}$$\leq$$v_{n}^{t}$$\leq$$h$), and $c_n$ is a label from class $\mathbb{C} = \{object, ground, sky\}$.

Each stixel class defines a different theoretical model for the disparity of each pixel $v$ along the stixel, $v_{n}^{b} \leq v \leq v_{n}^{t}$. This model is defined as a function $f$:

\begin{itemize}
\item The function for ground stixels is defined to match the disparity gradient of the ground surface: $f^{ground}(v) = \alpha(v_{horizon}-v)$ 
\item The function for sky stixels is zero (modeling very far away pixels): $f^{sky}(v) = 0$ 
\item The function for object stixels is supposed to have constant disparity, but this value depends on the particular object corresponding to the stixel. We model the function as $f_n^{object}(v) = f_n$, and this constant value is computed as the mean of the measured disparities of the considered stixel.
\end{itemize}

$L$$\in$$\mathbb{L}$ is an ordered list of $N$ consecutive, adjacent, and non-overlapped stixels $\{s_n\}, 1$$\leq$$n$$\leq$$N$$\leq$$h$, where $\mathbb{L}$ is the set of possible segmentations. Then, the stixel computation problem is formulated as a Maximum A Posteriori (MAP) estimation problem:
\begin{equation}
L^{*} = \arg\!\max_{L \in \mathbb{L}} P(L|D_i)
\label{eq:MAP}
\end{equation}

\subsection{Data \& Smoothness terms}

Applying the Bayes' theorem, the posterior probability $P(L|D_i)$ in Eq. \ref{eq:MAP} can be rewritten as $P(L|D_i) \approx P(D_i|L)P(L)$, where the first term, or data term, is the conditional probability of having the input column $D_i$ given a labeling $L$, and the second term, or smoothness term, is the prior probability of the configuration $L$. 

It is assumed that the data term can be computed independently for each pixel in the stixel and for all the stixels in the segmentation. This allows to express the data term with the following formula:

\begin{equation}
P(D_i|L) = \prod_{n=1}^{N} \prod_{v=v_{n}^{b}}^{v_{n}^{t}} P_{Di}(d_v|s_n,v)
\end{equation}

The data term $P_{Di}(d_v|s_n,v)$ models the probability for a single disparity measurement $d_v$ at row $v$ to belong to a given stixel $s_n$. 
A sensor model is used to estimate the likelihood (or cost) that the measured disparity data ($d_v$) matches the theoretical disparity model corresponding to a given stixel ($f(v)$). Following the proposal in \cite{PfeifferThesis}, we define this model as a combination of a Gaussian and a Uniform distribution.

\begin{equation}
P_{Di}(d_v|s_n,v) = \frac{p_{out}}{d_{range}}+\frac{1-p_{out}}{A_{norm}}e^{-\frac{1}{2}(\frac{d_v-f(v)}{\sigma^{c_n}(f,v)})^2}
\label{eq:sensor_model}
\end{equation}

The uniform distribution models the probability of the occurrence of an outlier ($p_{out}$ is the outlier rate), sometimes due to non-valid disparity measurements or by incorrectly matched pixels on the scene. $d_{range}$ is the total number of disparities considered in the input disparity map. The Gaussian distribution assesses the affinity of the measured disparity with the theoretical disparity function of the stixel. $A_{norm}$ is a normalization term, and $\sigma^{c_n}(f,v)$ is a sigmoid function that incorporates the noise model for the disparity measurement.

In order to avoid numerical problems with small magnitudes of the individual probabilities and to simplify Eq. \ref{eq:sensor_model}, the MAP estimation problem is expressed using the logarithm of the likelihoods instead of the actual likelihood. The optimization problem is then converted to a cost minimization problem. Following is the final equation to model the data term cost of a given pixel $v$ belonging to a given stixel $s_n$. The data term cost of a stixel is the aggregation of the costs of all of its pixels, $C_{data}(s_n) = \sum_{v=v_{n}^{b}}^{v_{n}^{t}} C_{data}(d_v|s_n,v)$
\begin{equation}
\begin{split}
C_{data}(d_v|s_n,v) = min(log(d_{range})-log(p_{out}), \\
log(A_{norm})+log(\sigma_{c_n}(f,v)\times\sqrt[]{2\pi})-log(1-p_{out}) \\
+\frac{1}{2\sigma_{c_n}^2(f,v)}(d_v-f(v))^2
\end{split}
\label{eq:log_sensor_model}
\end{equation}

The prior probability, or smoothness term, models the real-world constraints described at the beginning of this section, and is defined as a set of cost tables (log-likelihoods instead of actual probabilities). These constraints only consider the likelihood of the first stixel and the pairwise mutual dependencies of a pair of adjacent stixels $C_{prior}(s_n, s_{n-1})$. In our proposal we use the same model and parameters described in \cite{PfeifferThesis,Pfeiffer2011}.

\subsection{Solving Stixels with Dynamic Programming}
\label{subsec:solving_dp}

Dynamic Programming solves problems by breaking them down to simpler subproblems and storing the partial solutions on a memory structure. This way, when a given subproblem appears again, computation time is saved by retrieving the partial solution from the memory structure and not by solving the same subproblem repeatedly. 

The dynamic programming scheme is used for computing the stixel segmentation $L$= $\{s_n\}$ with minimum global cost for a column $D_i$. The global cost is composed of a data term $C_{data}(L)= \sum_{s_n\in L}C_{data}(s_n)$ and a smoothness term $C_{prior}(L)= \sum_{n=1}^{N}C_{prior}(s_n,s_{n-1})$. For that purpose, we need to express the optimization problem as a composition of smaller sub-problems.

In order to simplify our description we use a special notation to refer to the three different types of stixels considered in this work: $ob_{b}^{t} = \{v^{b}, v^{t}, object\}$, $gr_{b}^{t} = \{v^{b}, v^{t}, ground\}$, and $sk_{b}^{t} = \{v^{b}, v^{t}, sky\}$. Similarly, $OB^k$, $GR^k$, and $SK^k$ are defined as the minimum aggregated cost of the best segmentation of column $D_i$ from position $0$ to position $k$, both included, for three cases: each case corresponds to a segmentation ending on a stixel of the corresponding class. The stixel at the end of the segmentation associated with each minimum cost is denoted as $ob^k$, $gr^k$, and $sk^k$, respectively. We next show a recursive definition of the problem that can be solved by dynamic programming:

\begin{equation}
\begin{split}
OB^0 = C_{data}(ob_0^0)+C_{prior}(ob_0^0)\\
GR^0 = C_{data}(gr_0^0)+C_{prior}(gr_0^0)\\
SK^0 = C_{data}(sk_0^0)+C_{prior}(sk_0^0)
\end{split}
\label{eq:dyn_programming_zero}
\end{equation}

\begin{equation}
\begin{split}
OB^k = min \begin{cases} C_{data}(ob_0^k)+C_{prior}(ob_0^k)\\
C_{data}(ob_1^k)+C_{prior}(ob_1^k,ob^0)+ OB^0\\
C_{data}(ob_1^k)+C_{prior}(ob_1^k,gr^0)+ GR^0\\
C_{data}(ob_1^k)+C_{prior}(ob_1^k,sk^0)+ SK^0\\
...\\
C_{data}(ob_k^k)+C_{prior}(ob_k^k,ob^{k-1})+ OB^{k-1}\\
C_{data}(ob_k^k)+C_{prior}(ob_k^k,gr^{k-1})+ GR^{k-1}\\
C_{data}(ob_k^k)+C_{prior}(ob_k^k,sk^{k-1})+ SK^{k-1}
\end{cases}
\end{split}
\label{eq:dyn_programming}
\end{equation}

Eq. \ref{eq:dyn_programming_zero} represents the base case problem: segmenting a column of the single pixel at the bottom. Eq. \ref{eq:dyn_programming} indicates how to solve a problem of size $k$ using the solutions for smaller problems, computed so far. We only show the case for object stixels, but the other cases are solved similarly. All the possible object stixels ending at position $k$ (and starting at positions from $0$ to $k$) are connected with the last stixel of the minimal cost segmentations of the corresponding size, which were previously computed. Connections are evaluated for the three stixel classes using the smoothness term (prior model). 

All the partial solutions $OB^k$, $GR^k$, and $SK^k$, are stored in a cost table $C$ during the solving steps of the recurrent algorithm. As shown by Eq. \ref{eq:dyn_programming}, solving a sub-problem of size $k$ using the previous solutions for $j$$:$$0$$\leq$$j$$\leq$$k$, requires considering the $k$ possible positions of a cut between stixels and the 3 possible classifications of the stixels. Assuming that the number of different classes is constant (3 classes), the complexity of the stixel estimation problem for a single column is $\mathcal{O}(h \times h)$, and the complexity for the stixel segmentation of the whole disparity image is $\mathcal{O}(h^2\times\frac{w}{s})$.

Once the cost table $C$ is completely calculated, and in order to find the optimal configuration of stixels, a backtracking procedure has to be performed starting from the top row of $C$ and computing the successive minimum value $C_{min}^{h-1} = min(OB^{h-1}, GR^{h-1}, SK^{h-1})$. This task is sped up by using an index table that links each stixel to the next stixel with minimum cost, which is updated during the solving process.

\subsection{Using LUTs to optimize performance}
\label{subsec:algorithmic_optimizations}
The usage of Look-Up Tables (LUTs) reduces the amount of computation and memory accesses required to solve the problem, and assures that the algorithmic complexity is the one calculated so far. First, we review some optimizations presented in \cite{PfeifferThesis}, and then present a new one that provides good results both on CPU and GPU.

Computing the cost for each of the stixels considered when solving the problem using Eq. \ref{eq:log_sensor_model} is very expensive. But most of the terms in the equation do not depend on the input data and can be pre-computed. Accordingly, only the last term has to be actually computed for each disparity measurement ($d_v$). We must consider two different cases. 

The cost for pixels classified as $ground$ or $sky$ only depends on the current disparity of the pixel. Then, the response of the sensor model provided by Eq. \ref{eq:log_sensor_model} can be pre-computed for each of the disparities in the input column, and stored in the LUT. Therefore, the total computation complexity for computing and storing the cost for $ground$ and $sky$ is $\mathcal{O}(h\times\frac{w}{s})$.

The cost of a pixel classified as $object$, though, depends not only on the current disparity of the pixel but also on the mean disparity of the segment. In order to limit the total amount of possible input combinations and to reduce the complexity of pre-computing all the corresponding cost values, we round the mean disparities into integer values. This approach provides satisfactory quality results while reducing the number of pre-computed values to all the possible combinations of the $h$ disparity values on each input column and the $d_{range}$ possible average disparities, which account for a total of $h$$\times$$d_{range}$ values.

The algorithmic improvement with more impact on performance comes from the use of prefix sums \cite{Blelloch89} to reduce the total amount of operations needed to calculate the total cost of the pixels of a stixel. The prefix sum of a vector of numbers is a new vector that holds the accumulated cost corresponding to the first k pixels. Prefix sums extended to 2D or 3D matrices are known in the field of image processing as integral images or summed-area tables \cite{ViolaJones}.

The LUTs described so far contain the prefix sum of the costs corresponding to each pixel in the input column. Then, calculating the cost of a stixel $s_n = \{v_{n}^{b}, v_{n}^{t}, c_n\}$ is done in constant time just by subtracting two numbers in the table, independently of the size of the stixel. The LUTs of the pre-computed cost for $ground$ or $sky$ stixels are indexed just by the positions of the first and last pixels of the stixel: $C_{data}(s_n) = LUT_{c_n}[v_n^t]-LUT_{c_n}[v_n^{b-1}]$. The LUT of the pre-computed cost for $object$ stixels is indexed by the positions of the stixel but also by the average disparity of the stixel ($f_n$): $C_{data}(s_n) = LUT_{object}[f_n][v_n^t] - LUT_{object}[f_n][v_n^b-1]$. Again, the average disparity of a given stixel, $f_n$, is computed in constant time using a pre-computed prefix sum of the disparities of the pixels in the processed column.

In our design we propose using a new LUT containing the pre-computed costs for all possible pairs of pixel disparity and mean disparity (a 2D matrix of $d_{range} \times d_{range}$ elements). Since the contents of this table are independent on the input data, the table can be computed off-line and avoids any computation from Eq. \ref{eq:log_sensor_model} to be executed during the normal process of stixel estimation. We have experimentally verified that this new LUT improves the performance both on CPU and on GPU.

To summarize, some LUTs are computed off-line, while the LUTs containing prefix sums must be computed for each new disparity image. The higher computational work of generating the LUTs  corresponds to the creation of the two-dimensional $object$ LUTs, with a complexity of $\mathcal{O}(h \times d_{range} \times \frac{w}{s})$ operations per input image. This complexity is comparable to the complexity of the dynamic programming step as long as $h$ is higher or equal to $d_{range}$, which is often the case.

\section{GPU architecture and performance}
\label{sec:gpu}
GPUs are massively-parallel devices containing tens of throughput-oriented processing units called \emph{streaming multiprocessors} (SMs). Memory and compute operations are executed as vector (SIMD) instructions and are highly pipelined in order to save energy and transistor budged. SMs can execute several vector instructions per cycle, selected from multiple independent execution flows: the higher the available thread-level parallelism the better the pipeline utilization.

The CUDA programming model allows defining a massive number of potentially concurrent execution instances (called $threads$) of the same program code. A unique two-level identifier \textless $ThrId$, $CTAid$\textgreater $ $ is used to specialize each thread for a particular data and/or function. A $CTA$ ($Cooperative$ $Thread$ $Array$) comprises all the threads with the same $CTAid$, which run simultaneously and until completion in the same SM, and can share a fast but limited memory space: the so-called $Shared Memory$. $Warps$ are groups of threads with consecutive $ThrId$s in the same CTA that are mapped by the compiler to vector instructions and, therefore, advance their execution in a lockstep synchronous way. The warps belonging to the same CTA can synchronize using an explicit barrier instruction. Each thread has its own private $Local Memory$ space (commonly assigned to registers by the compiler), while a large space of $Global Memory$ is public to all execution instances (mapped into a large-capacity but long-latency device memory, which is accelerated using a two-level hierarchy of cache memories).

The parallelization scheme of an algorithm and the data layout determine the available parallelism at the instruction and thread level (required for achieving full resource usage) and the memory access pattern. GPUs achieve efficient memory performance when the set of addresses generated by a warp refer to consecutive positions that can be \emph{coalesced} into a single, wider memory transaction. Since the bandwidth of the device memory can be a performance bottleneck, an efficient CUDA code should promote data reuse on shared memory and registers.

\section{Massive Parallelization}
\label{sec:parallelization}

This section describes and discusses the parallelization schemes and data layouts used for the algorithms involved in the stixel computation pipeline.

\subsection{Column Reduction and Transpose}

\begin{figure}
\includegraphics[width=\linewidth]{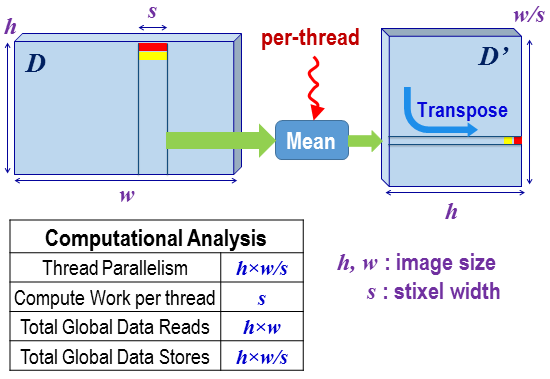}
\caption{Column Reduction and Transpose of input Disparity image: parallel scheme and computational analysis}
\label{fig:preprocessing}
\end{figure}

Stixels are computed on columns of width $s$ and height $h$. The first step in the pipeline is to reduce the width of each column into a single pixel by replacing the disparities of $s$ consecutive pixels in the same row by their average. 
The input disparity image arranges pixels into consecutive rows of memory (row-wise), but this is not the appropriate data layout for the tasks that are required later, where information is processed in columns. 
Therefore, we fuse a transpose operation with the column reduction operation described so far into the same algorithmic step. The scheme of the operation is shown in Fig. \ref{fig:preprocessing}.

The data access pattern of the algorithm exhibits no data reuse and, therefore, thread cooperation is not required. As shown in Fig. \ref{fig:preprocessing}, work is distributed by assigning to each thread the task of reducing the disparities of $s$ consecutive pixels into a single output disparity value, which must be stored in a transposed position. Each thread reads $s$ consecutive input disparities, computes their mean, and writes one result. We further assign the consecutive threads of a warp to consecutive output positions, so that writes are  coalesced and write performance is maximized. Reads, though, are not coalesced (a typical situation on transpose operations) and will perform sub-optimally. We improve read performance by reading consecutive pixels in groups of 4 or 2 when possible. A collaborative read strategy using Shared Memory would slightly improve performance but, since this processing step represents a very small amount of time on the pipeline ($\leq 0.5\%$), we preferred a simple solution. 

\subsection{Pre-computation of $LUT_{object}$}

As explained in subsection \ref{subsec:algorithmic_optimizations}, a specific look-up table for the $object$ data-term ($LUT_{object}$) has to be generated for each input column $D_i$, for a total of $\mathcal{O}(h \times d_{range}\times \frac{w}{s})$ output values. Since $LUT_{object}$ is too large to fit into Shared Memory, it must be written to Global Memory, and then we can isolate this task from the whole processing pipeline without losing performance.

\begin{figure}
\includegraphics[width=\linewidth]{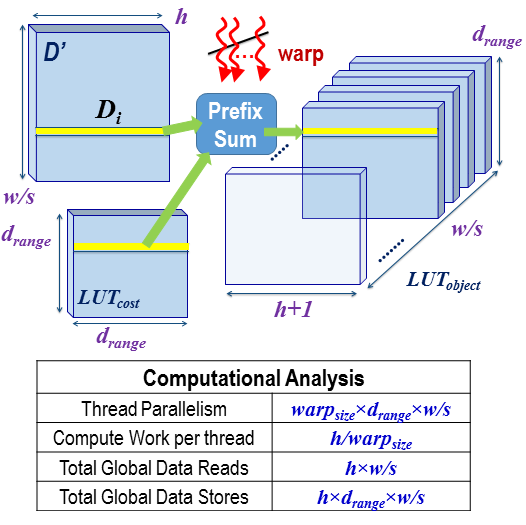}
\caption{Pre-computation of $LUT_{object}$: parallel scheme and computational analysis}
\label{fig:precomputation}
\end{figure}

As shown in Fig. \ref{fig:precomputation}, work is distributed by assigning to a single warp the task of generating one row of the $LUT_{object}$ matrix corresponding to a single input column $D_i$. Warps in the same CTA cooperate by reading the input column into Shared Memory. Then, each warp computes the prefix sum of the cost vector corresponding to one row in the LUT. The warp iterates on the $h$ elements of $D_i$, processing $warp_{size}=32$ elements on each iteration step. Data read from $LUT_{cost}$ is used straightforwardly to compute the prefix sum directly into registers (Local Memory), using register-to-register $shuffle$ instructions, and affording memory reads and writes. No explicit synchronization is required when operating warp-wise, as shown by \cite{harris2007parallel}.

The performance bottleneck of this stage, which represents less than $5\%$ of the time of the whole pipeline, is the write bandwidth to the external device memory.

\subsection{Dynamic Programming (DP) stage}
This is the most time-consuming ($\geq 95\%$) processing step, and also the most elusive for massive parallelization. Each input column $D_i$ can be processed independently to generate a list of estimated stixels, but this amount of parallelism is not enough to efficiently exploit current GPUs for the image sizes considered in most real-world applications. The challenge is to extract fine-grain parallelism inside the DP task corresponding to each input column.

\begin{figure}
\includegraphics[width=\linewidth]{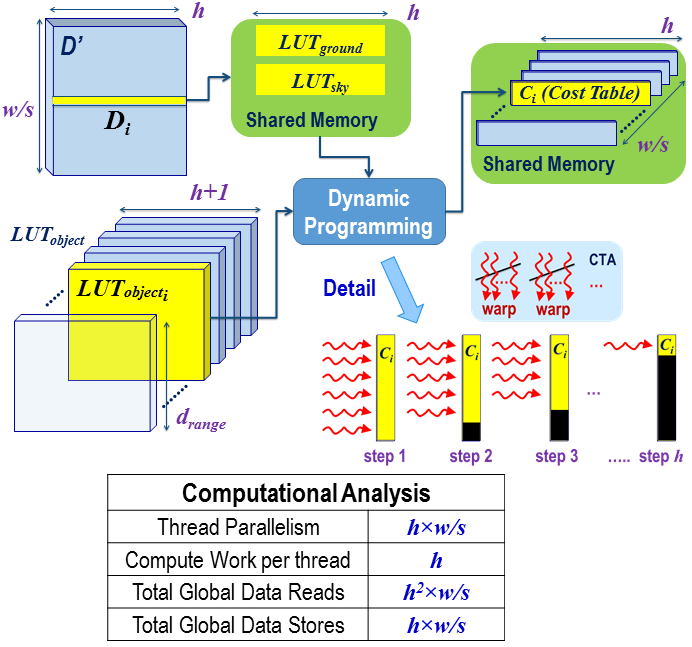}
\caption{Dynamic Programming (DP) stage: parallel scheme and computational analysis}
\label{fig:stixels_computation}
\end{figure}

Processing column $D_i$ involves multiple repeated reads to all the corresponding LUTs, and reading and updating the contents of the corresponding cost table $C_i$, as shown (yellow) in Fig. \ref{fig:stixels_computation}.
In order to improve data access performance we promote data reuse by assigning a separate Cooperative Thread Array (CTA) of $h$ threads to each DP task. Before performing the actual DP solving process, threads cooperate to copy some general LUTs from Global Memory to Shared Memory and to compute the prefix sums of $LUT_{ground}$, $LUT_{sky}$ and $D_i$ into Shared Memory using \cite{harris2007parallel}. $LUT_{object_i}$ is the only data structure that does not fit into Shared Memory and must be accessed from Global Memory.
 
The DP recurrence shown in Eq. \ref{eq:dyn_programming} formulates how to calculate the minimum cost of a sub-problem with $k$ pixels using the results computed for smaller sub-problems. We distribute the DP solving task by making each CTA thread responsible of computing the minimum cost for each problem size $k$ ($0$$\leq$$k$$<$$h$). Two issues are derived from Eq. \ref{eq:dyn_programming} that complicate parallel execution: (1) the work assigned to each thread is not well balanced, since it is proportional to $k$ (see the sequence of steps depicted in Fig. \ref{fig:stixels_computation}); and (2) there are data dependencies that must be preserved using synchronization operations. The cost table $C_i$ (in Shared Memory) is used to communicate partial results between threads, and barrier synchronization is used to enforce dependencies among the consecutive DP steps. Also, every step of the DP solving process decreases the number of active parallel threads.

Synchronization barriers between recurrent steps, reduced warp parallelism in the CTA as the recurrence loop advances (an average of half the warps are active on each CTA), and moderate warp divergence (an average of half the threads are active on the last warp), prevent using the available computation resources efficiently, making performance latency-bounded.

\subsection{Backtracking}

\begin{figure}
\includegraphics[width=\linewidth]{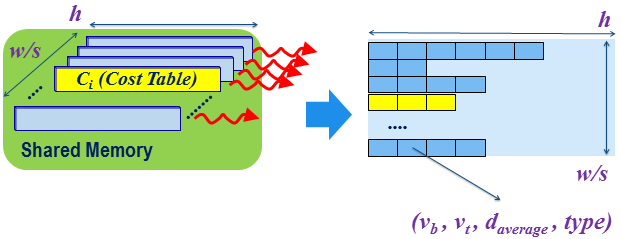}
\caption{Backtracking: Parallel scheme}
\label{fig:backtracking}
\end{figure}

The backtracking step is an inherently sequential process (for each column). As described in subsection \ref{subsec:solving_dp}, it navigates back on an index table created during the DP solving stage (not shown in the figures) and produces a list of stixels with the optimal configuration for the column (see Fig. \ref{fig:backtracking}).

The lack of parallelism in this final step seems to discourage a GPU implementation, but we have found that the time to transfer the resulting index tables to the CPU, or even from Shared Memory to Global Memory, is higher than the time to perform the task on the GPU (less than $0.5\%$ of the overall execution time). Therefore, we fuse this computing stage with the DP solving stage described in the last subsection, and the last active thread in a CTA is responsible of generating the final output. In order to overcome the problem of handling variable-size lists of stixels, we pre-allocate a fixed amount of Global Memory for each list. 

\section{Results}
\label{sec:results}

This section assesses the quality results and the performance of our proposal. A previous concern is to verify that our proposal conforms to the algorithm defined so far,  and adopted from \cite{Pfeiffer2011}. For that purpose, we used both synthetic and real data. Stereo images generated using SYNTHIA \cite{RosCVPR16} (like the one shown in Fig. \ref{fig:overview}) provide examples with exact disparity maps and free space identification\footnote{We thank the authors of SYNTHIA for providing us with the data}. All the experiments using images including cars, pedestrians, trees, and traffic signals, provided the expected results.

We have used the manually-labeled data-set provided by  \cite{pfeiffer2013exploiting} for a preliminary evaluation of our implementation. We have selected $\approx$ 1500 stereo images from the data-subset corresponding to good weather conditions, since for those images we can generate acceptably good stereo disparity map estimations at real-time with \cite{HCE2016}. We use the following metrics for comparing the quality of our stixel estimation with the provided Ground Truth (GT) stixel results:
\begin{itemize}  
\item Detection Rate: we consider that a GT stixel has been detected if the ratio of pixels that intersect with an estimated object stixel with respect to the size of the GT stixel is higher than 0.5. We show the proportion of detected versus total number of GT stixels.
\item False Positives: a stixel classified as an object is considered a false positive when more than 30 pixels are inside the free space determined by the GT stixels.
\end{itemize}

\begin{table}
\begin{center}
\begin{tabular}{|l|c|}
\hline
Detection Rate & 88.7 \% \\
\% Image Pairs with False Positives & 2.14 \% \\ 
Total number of False Positives & 155 \\ 
\hline
\end{tabular}
\end{center}
\caption{Quality results obtained using a publicly available data-set composed of 1495 image pairs}
\label{tab:quality}
\end{table}

\begin{figure}
\includegraphics[width=\linewidth]{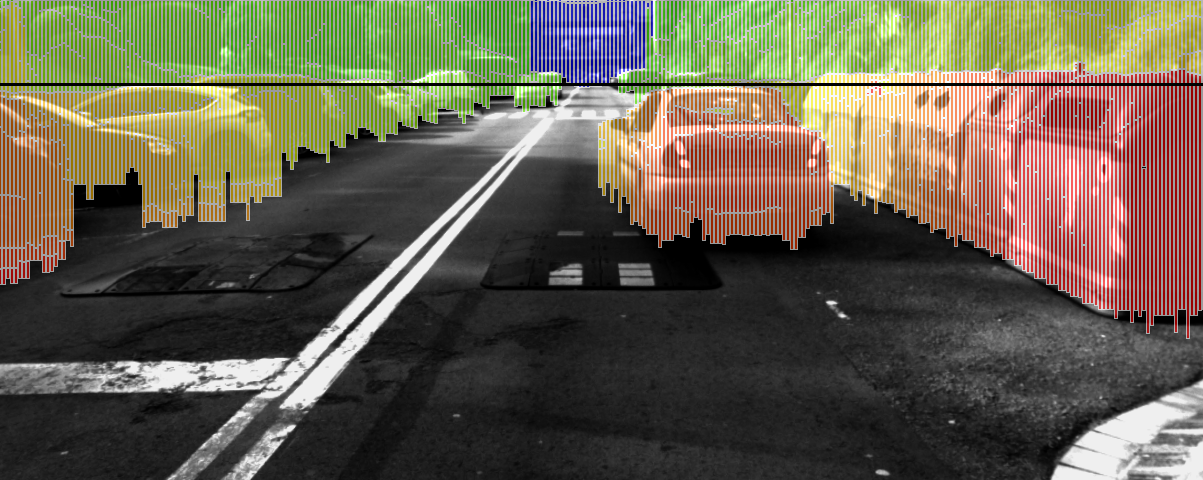}
\caption{Example of Stixel World estimation. Sky stixels are represented on blue, object stixels are represented on green-to-red (read= close, green= far), and ground stixels are transparent.}
\label{fig:stixels_example}
\end{figure}

Table \ref{tab:quality} shows the quality results obtained, which indicate that our proposal provides similar results as \cite{Pfeiffer2011}. A visual example of the stixel configuration computed by our proposal can be seen in Fig. \ref{fig:stixels_example}. We were not able to compare our proposal with the original CPU implementation, since we could not obtain the stereo disparity maps used on that work and the metrics were not properly described.

We have also used multiple images of different sizes, both real and synthetic, for the performance analysis. Our main goal was to evaluate performance on a NVIDIA Tegra X1 processor integrating 8 ARM cores and 2 Maxwell Streaming Multiprocessors (SMs), and with a Thermal Design Power (TDP) of 10 Watts. For comparison purpose, we have also measured performance on a high-end NVIDIA Titan X (Maxwell), with 24 Maxwell SMs and a TDP of 250W. We discard the time for CPU-GPU data transfers based on the fact that stixel estimation is just one stage in a general computation pipeline, which receives a disparity map from an stereo computation stage and provides a list of stixels for further post-processing.

\begin{figure}
\includegraphics[width=\linewidth]{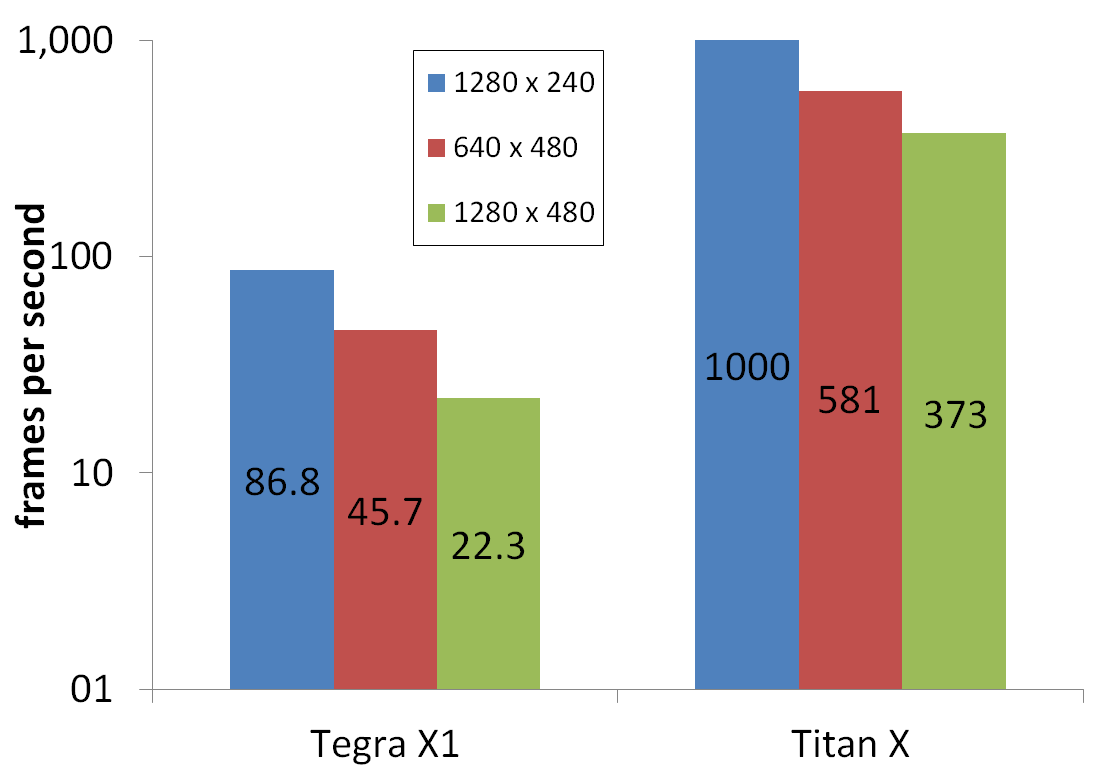}
\caption{Frames per second obtained for different image resolutions ($w$$\times$$h$) on NVIDIA Tegra X1 and Titan X, for $s=5$.} 
\label{fig:fps}
\end{figure}

\begin{figure}
\includegraphics[width=\linewidth]{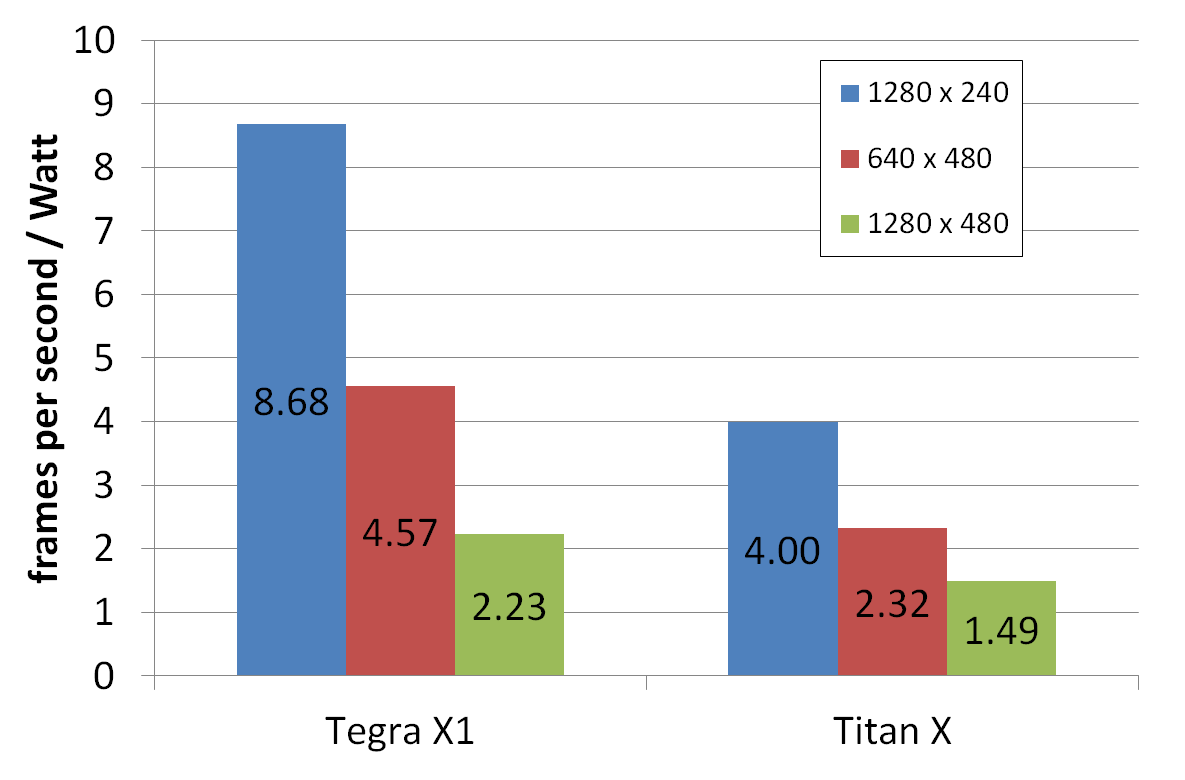}
\caption{Energy efficiency (FPS/Watt) for different image resolutions ($w$$\times$$h$) on NVIDIA Tegra X1 and Titan X, for $s=5$.}
\label{fig:energy_efficiency}
\end{figure}

Fig. \ref{fig:fps} and \ref{fig:energy_efficiency} show, respectively, the performance throughput (frames per second, or fps) and the performance per watt (fps/W) on both GPU systems and also for different image resolutions. The high-end GPU always provides more than 11 times the performance of the embedded GPU (as expected by the difference in number of SMs), but the latter offers between 1.5 and 2 times more performance per Watt. It is remarkable that real-time rates (22.3 fps) are achieved by the Tegra X1 with $1280$$\times$$480$ resolution. Also,  a high-end Titan X achieves very high performance, e.g. around 373 fps with $1280$$\times$$480$ resolution.

The algorithm implemented by \cite{PfeifferThesis} reaches 13.3 frames per second on a multi-core CPU (Core i7 980X, $6$$\times$$3.4$ Ghz, 6 GB of RAM) for a input disparity image of $1024$$\times$$440$ px and a stixel width of 5 px. Our implementation reaches 26 fps on a Tegra X1 for that resolution (and 413 fps on a Titan X). Therefore, performance is improved almost 2 times with respect to \cite{PfeifferThesis}, while the performance per watt ratio is 25 times better. As expected, the algorithmic complexity makes execution time to grow linearly with the image width but quadratically with the image height.

\section{Conclusions}
\label{sec:conclusions}
This paper has described and assessed the performance of the -to the best of our knowledge- first GPU-accelerated implementation of stixel estimation.  Results have shown that our proposal achieves real-time performance for realistic problem sizes, proving that the low-power envelope and remarkable performance of embedded CPU-GPU hybrid systems make them good target platforms for most real-time video processing tasks, paving the way for more complex and larger applications.

The core of stixel estimation involves a Maximum A Posteriori (MAP) probabilistic formulation that is solved using a Dynamic Programming (DP) scheme. We have proposed a parallel scheme and data layout for this computational pattern that follows general optimization rules based on a simple GPU performance model, and is then expected to scale gracefully on the forthcoming GPU architectures, like Pascal. Our proposal could be applied to similar DP computational patterns.

Since performance and low consumption are always welcome, for example to handle more and larger input images, we will explore alternative algorithmic strategies to further improve performance while maintaining good quality. Also, we will incorporate tracking and clustering on the GPU-accelerated pipeline, which will open new opportunities for improving stixel estimation quality.



{\small
\bibliographystyle{ieee}
\bibliography{egbib}
}

\end{document}